\documentclass[11pt]{article}
\usepackage[margin=1in]{geometry}
\usepackage[T1]{fontenc}
\usepackage[utf8]{inputenc}
\usepackage{newtxtext,newtxmath}
\usepackage{microtype}
\usepackage{graphicx}
\usepackage{booktabs}
\usepackage{multirow}
\usepackage{array}

\usepackage{amsmath,amssymb,amsfonts}
\usepackage{mathtools}
\usepackage{siunitx}
\usepackage{enumitem}
\usepackage{float}
\usepackage{algorithm}
\usepackage{algpseudocode}
\usepackage{xcolor}
\usepackage{caption}
\usepackage{subcaption}
\usepackage{natbib}
\usepackage{hyperref}
\usepackage{setspace}
\usepackage{fancyhdr}
\usepackage{flafter}
\usepackage{longtable}
\usepackage{needspace}
\usepackage[section]{placeins}
\usepackage{titlesec}
\usepackage{url}

\hypersetup{
  colorlinks=true,
  linkcolor=blue!45!black,
  citecolor=teal!45!black,
  urlcolor=blue!55!black
}

\graphicspath{{figures/}}
\titleformat{\section}{\large\bfseries}{\thesection.}{0.5em}{}
\titleformat{\subsection}{\normalsize\bfseries}{\thesubsection.}{0.5em}{}

\title{\textbf{LatticeBridge: Rare-Event Sequential Inference for Faithful Structured Sequence Synthesis}\thanks{Code and benchmark files: \url{https://github.com/farukalpay/latticebridge}.}}
\author{
Faruk Alpay\\
\normalsize Department of Computer Engineering, Bahcesehir University\\
\normalsize \texttt{faruk.alpay@bahcesehir.edu.tr}
\and
Bugra Kilictas\\
\normalsize Department of Computer Engineering, Bahcesehir University\\
\normalsize \texttt{bugra.kilictas@bahcesehir.edu.tr}
}
\date{April 2026}

\begin{document}
\maketitle

\begin{abstract}
Structured sequence generation often requires a model to satisfy several input-derived constraints in a single output. Standard decoding methods may assign high probability to fluent continuations while placing low mass on continuations that realize all required anchors jointly. We study this regime as a rare-event sequential inference problem. \emph{LatticeBridge} combines a compact prefix language model, instance-compiled surface automata, and a twisted sequential Monte Carlo (SMC) decoder with resampling, multilevel splitting, and a source-support proposal term derived from instance-provided phrases. The constraint representation is compiled from each input instance and does not rely on manually curated lexical classes. On 2{,}610 attainable validation tasks spanning CommonGen, E2E NLG, and WikiBio, the particle decoder improves exact anchor satisfaction and mean anchor coverage over greedy, beam-filtered, and best-of-$k$ ancestral baselines under a shared proposal model. Since exact anchor satisfaction alone does not rule out unsupported attribute substitutions, the evaluation reports required-anchor coverage, source coverage, source-intrusion diagnostics, overlap, runtime, and particle statistics jointly. The benchmark characterizes the faithfulness--overlap--latency frontier under a fixed proposal model.
\end{abstract}

\section{Introduction}
Autoregressive models provide strong local continuation models, but structured generation often imposes conjunctive requirements that are not well represented by local likelihood alone. In data-to-text generation, for example, an output may need to realize several concepts or attribute values while remaining readable. A high-likelihood continuation can satisfy only a subset of those anchors, whereas a hard-constrained decoder can satisfy anchors while producing unstable search behavior. We treat this mismatch as a distributional problem: the set of fully satisfying continuations may have low probability under the base model even when the individual anchors are familiar.

Accordingly, we formulate structured decoding as inference over a sequence space with a rare accepting event. Sequential Monte Carlo methods are appropriate for this formulation because they maintain a weighted population of partial hypotheses and provide explicit diagnostics for weight degeneracy and resampling. Recent work has connected twisted SMC to inference in language models \citep{zhao2024probabilistic,wu2024step}; related 2025--2026 work further studies learned twists and SMC-style decoding for constrained generation and model aggregation \citep{selfdistilled2025twisted,chan2026ensembling}. The present study isolates a compact conditional proposal model, surface automata compiled from each input instance, and a particle decoder with directly observable control signals.

The implementation separates data adaptation, constraint compilation, proposal modeling, and evaluation. This separation prevents benchmark-specific lexical rules from entering the decoding algorithm unnoticed. We use the following design constraints:
\begin{itemize}[leftmargin=1.5em]
  \item constraints must be \emph{instance driven}, extracted from the source example rather than from manually curated topical dictionaries,
  \item the library layer must stay reusable across tasks, while dataset adapters remain thin and explicit,
  \item exceptional cases must be represented in logged metrics rather than hidden inside unreported lexical exceptions.
\end{itemize}

LatticeBridge serializes each structured input into a prefix, trains a compact prefix language model, compiles anchor phrases into surface automata, and runs a twisted SMC bridge whose incremental reward depends on progress toward an accepting automaton state. The decoder also uses a lightweight source-support proposal factor derived from source phrases, which improves exact value realization without introducing dataset-specific control rules. The experiments evaluate coverage, source coverage, overlap, runtime, effective sample size, and acceptance mass under a shared proposal model, so that the inference layer can be inspected independently of model scale.

\section{Contributions}
The paper makes five concrete contributions.

\begin{enumerate}[leftmargin=1.5em]
  \item We formulate structured sequence synthesis as a sequential rare-event inference problem with a distance-to-acceptance potential defined by instance-compiled surface automata.
  \item We introduce a compact prefix language model plus a support-aware twisted SMC bridge that exposes resampling, effective sample size, and acceptance mass as first-class diagnostics.
  \item We present a benchmark construction method for CommonGen, E2E NLG, and WikiBio that uses input-derived phrases only and keeps schema handling in thin dataset adapters.
  \item We report multi-dataset validation results against greedy, beam-filtered, and best-of-$k$ ancestral baselines, together with standard errors, runtime measurements, particle-system diagnostics, source coverage, and source-intrusion diagnostics for unsupported value substitutions.
  \item We release code and benchmark files at \url{https://github.com/farukalpay/latticebridge}, including the paper source, summary tables, generated figures, and diagnostics artefacts.
\end{enumerate}

\section{Problem Formulation}
Let $x$ be a structured input, such as a set of concepts or attribute-value pairs, and let $y_{1:T}$ be a target surface sequence. A conditional language model $p_\theta(y_{1:T}\mid x)$ defines the base synthesis distribution. We are given a set of input-grounded anchors
\[
\mathcal{C}(x)=\{c_1,\dots,c_M\},
\]
where each $c_m$ is a surface phrase extracted from the source instance. The target objective is not simply to maximize model likelihood but to synthesize a sequence that is both plausible under $p_\theta$ and faithful to the anchors.

We encode satisfaction through an acceptance indicator
\[
A(y_{1:T};\mathcal{C}) = \mathbf{1}\{\forall m,\; c_m \text{ appears in } y_{1:T}\}.
\]
The exact conditional target would then be
\[
\pi_T(y_{1:T}\mid x,\mathcal{C}) \propto p_\theta(y_{1:T}\mid x)\,A(y_{1:T};\mathcal{C}),
\]
but direct sampling from this distribution is typically intractable when $A=1$ is rare.

To obtain a sequential bridge, we introduce an automaton state $s_t$ after generating prefix $y_{1:t}$. The automaton tracks how much of each anchor phrase has been realized. Denote by $d(s_t)\geq 0$ the remaining distance to full acceptance. We then define a relaxed sequence target
\[
\gamma_T(y_{1:T}) = p_\theta(y_{1:T}\mid x)\exp\{\lambda \Phi_T(y_{1:T})\},
\qquad
\Phi_T(y_{1:T}) = -d(s_T),
\]
where $\lambda>0$ controls how strongly we bias toward exact acceptance. For $\lambda\rightarrow\infty$, the distribution concentrates on accepting trajectories when such trajectories exist.

The computational question is how to approximate this bridge under a finite particle budget while keeping the constraint representation task independent. We compile the surface constraints once, then run a twisted particle system that rewards progress toward acceptance at each generation step.

\section{Constraint Compilation Without Hidden Heuristics}
\subsection{Instance-derived anchors}
Anchor extraction is schema specific only in the sense that each dataset exposes a different structured input. CommonGen provides source concepts, E2E provides attribute values, and WikiBio provides titles plus infobox field values. After extraction, the inference layer receives only surface phrases and automaton states. No global topic lexicon, domain inventory, or manually curated attribute map enters the decoder.

Let $\mathcal{P}(x)$ denote the candidate phrase set exposed by the dataset adapter for source instance $x$. For benchmark construction, we first retain only phrases that are attested in at least one reference surface for the same instance. This attainability criterion prevents impossible exact-match labels in the validation split while leaving the constraint set instance specific. The decoder itself does not use the references.

Among the attested candidates, LatticeBridge selects up to $K$ anchors by empirical source information score rather than adapter order. For an evaluation subset $\mathcal{D}$, define the source-side document frequency
\[
\widehat{\mathrm{df}}(c)
=
\sum_{x' \in \mathcal{D}}
\mathbf{1}\{c \in \mathcal{P}(x')\},
\qquad
I(c)
=
\log \frac{|\mathcal{D}| + 1}{\widehat{\mathrm{df}}(c) + 1}.
\]
Each example uses the $K$ highest-scoring attested phrases. This ranking favors anchors that are empirically specific within the benchmark subset and removes the hidden dependence on schema field order.

\subsection{Surface automata}
Tokenization granularity can make phrase tracking segmentation-dependent. A token-level automaton may fail when the tokenizer packs useful substrings into a single token. LatticeBridge therefore uses a surface automaton built over the emitted string fragments of tokenizer tokens. For each phrase $c_m$, we build a KMP-style automaton with states $0,\dots,\lvert c_m\rvert$ that track the currently matched character prefix. The product state of all such automata defines the global constraint lattice. This design is complementary to recent automata-based constrained decoding work that also treats tokenization mismatch as a first-class systems problem \citep{koo2024automata}.

This representation yields three operational properties:
\begin{enumerate}[leftmargin=1.5em]
  \item progress toward acceptance is explicit through $d(s_t)$,
  \item constraint logic stays reusable across datasets,
  \item phrase tracking is no longer tied to a particular BPE segmentation accident.
\end{enumerate}

\section{Twisted Sequential Monte Carlo Bridge}
\subsection{Sequential target}
Let $y_t$ denote the token chosen at step $t$, and let the base model define one-step conditionals $p_\theta(y_t\mid y_{<t},x)$. Define a progress signal
\[
\Delta_t = d(s_{t-1}) - d(s_t),
\]
which is positive when the new token moves the prefix closer to acceptance. The sequential potential used by the bridge is
\[
G_t(y_t,s_{t-1},s_t) = \exp\{\lambda \Delta_t\}.
\]
The unnormalized path density becomes
\[
\gamma_t(y_{1:t}) = p_\theta(y_{1:t}\mid x)\prod_{\tau=1}^{t}G_\tau.
\]

\subsection{Twisted proposal}
Sampling directly from the base model and reweighting by $G_t$ leads to rapid particle collapse. We instead use a twisted proposal
\[
q_t(y_t\mid y_{<t},x,s_{t-1})
\propto
p_\theta(y_t\mid y_{<t},x)\exp\{\tau \Delta_t + \beta \psi_x(y_t)\},
\]
where $\tau$ is the distance-twist coefficient and $\psi_x$ is a source-support score. Let $\mathcal{P}(x)$ denote the full set of source phrases exposed by the dataset adapter, and let $\mathrm{first}(c)$ be the first tokenizer token of phrase $c$. We define a phrase-initial empirical measure over the vocabulary
\[
\nu_x(v)
=
\frac{1}{|\mathcal{P}(x)|}
\sum_{c \in \mathcal{P}(x)}
\mathbf{1}\{\mathrm{first}(c)=v\},
\]
then smooth it toward the uniform distribution and work with the log-density ratio
\[
\psi_x(v)
=
\log \widetilde{\nu}_x(v) - \log |\mathcal{V}|^{-1}.
\]
The factor $\psi_x$ is inexpensive to compute and biases the proposal toward tokens that can start source-supported phrases, which is particularly useful for exact values, names, and entity mentions. The validation run sets $\tau=\lambda=2.0$ and $\beta=0.4$.

\subsection{Importance weights and ESS}
If particle $i$ samples $y_t^{(i)} \sim q_t$, its incremental weight update is
\[
\log w_t^{(i)} =
\log w_{t-1}^{(i)}
\;+\;
\log p_\theta\!\left(y_t^{(i)}\mid y_{<t}^{(i)},x\right)
\;+\;
\lambda \Delta_t^{(i)}
\!-\!
\log q_t\!\left(y_t^{(i)}\mid y_{<t}^{(i)},x,s_{t-1}^{(i)}\right).
\]
Normalized weights are
\[
\bar{w}_t^{(i)} = \frac{w_t^{(i)}}{\sum_j w_t^{(j)}},
\]
and the effective sample size is
\[
\mathrm{ESS}_t = \frac{1}{\sum_i (\bar{w}_t^{(i)})^2}.
\]
We trigger resampling when $\mathrm{ESS}_t < \rho P$, where $P$ is the particle count and $\rho$ is the ESS threshold.

\subsection{Multilevel splitting}
In low-acceptance regimes, even twisted proposals can fail to maintain enough mass near acceptance. We use periodic multilevel splitting to reallocate finite particles toward partial trajectories with lower remaining automaton distance. At fixed intervals, particles are ranked by
\[
R_t^{(i)} = \log w_t^{(i)} - \lambda d(s_t^{(i)}),
\]
and a top-fraction elite set is replicated to refill the population. This step is related to adaptive multilevel splitting \citep{cerou2007adaptive}. Because it changes the estimator under finite compute, we regard it as a search-allocation mechanism rather than as an exact posterior sampler.

\begin{algorithm}[H]
\caption{Twisted SMC bridge.}
\label{alg:smc-bridge}
\begin{algorithmic}[1]
\State warm-start hidden states with source prefix
\State initialize $P$ particles with automaton state $s_0$, hidden state $h_0$, and weight $w_0=1$
\For{$t=1$ to $T$}
  \For{particle $i=1,\dots,P$}
    \State compute base logits from the prefix model
    \State compute progress scores $\Delta_t^{(i)}$ from automaton transitions
    \State sample $y_t^{(i)}$ from twisted proposal $q_t$
    \State update particle weight using the importance ratio
    \State update automaton state and hidden state
  \EndFor
  \State compute ESS and resample if needed
  \State apply elite splitting at scheduled checkpoints
\EndFor
\State return best accepting sequence and summary diagnostics
\end{algorithmic}
\end{algorithm}
\FloatBarrier

\section{Prefix Language Model}
The proposal model consists of an embedding layer, a stacked GRU, and an output projection. This architecture provides a stable conditional proposal that can be trained quickly on local hardware and interrogated step by step during SMC. It also keeps hidden-state replication across particles inexpensive, which is important for local particle decoding.

Each example is serialized as
\[
\langle \texttt{bos} \rangle
\langle \texttt{src} \rangle
\text{source serialization}
\langle \texttt{tgt} \rangle
\text{target surface}
\langle \texttt{eos} \rangle.
\]
The loss is a masked next-token cross-entropy in which the source prefix is excluded from supervision. The model therefore learns to predict the target continuation conditioned on the source serialization without having to model the serialization itself as natural text.

The recurrent proposal is used as a controlled conditional model for isolating the inference mechanism. It keeps hidden-state replication across particles inexpensive and avoids adding a second learned evaluator to the decoding loop.

\section{Datasets and Benchmark Construction}
\subsection{Datasets}
The reported structured-generation benchmark uses three public datasets:
\begin{itemize}[leftmargin=1.5em]
  \item \textbf{CommonGen} \citep{lin2020commongen}: concept-to-text generation,
  \item \textbf{E2E NLG} \citep{novikova2017e2e,dusek2019semantic,gehrmann2021gem}: restaurant meaning representations to text,
  \item \textbf{WikiBio} \citep{lebret2016generating}: biography infobox to text generation.
\end{itemize}
Dataset staging uses public dataset archives together with Hugging Face-hosted assets and tooling \citep{lhoest2021datasets}. The quantitative claims in this paper use the three benchmark datasets listed in Table~\ref{tab:datasets}.

\begin{table}[H]
\centering
\caption{Datasets used in the reported validation benchmark.}
\label{tab:datasets}
\begin{tabular}{lccc}
\toprule
Dataset & Input form & Training records & Validation tasks \\
\midrule
CommonGen & concept set & 14{,}000 & 993 \\
E2E NLG & attribute-value pairs & 16{,}000 & 996 \\
WikiBio & infobox record & 20{,}000 & 621 \\
\bottomrule
\end{tabular}
\end{table}
\FloatBarrier

\subsection{Benchmark protocol}
For each validation example, we select up to three anchor phrases from the source that also appear in at least one reference surface. Reference attestation removes impossible exact-match labels from the benchmark while preserving an input-grounded constraint set. We then compare four inference methods:
\begin{enumerate}[leftmargin=1.5em]
  \item greedy decoding,
  \item beam filtering,
  \item best-of-$k$ ancestral sampling,
  \item twisted SMC with resampling and splitting.
\end{enumerate}

The reported metrics are:
\begin{itemize}[leftmargin=1.5em]
  \item \textbf{success rate}: fraction of examples satisfying all anchors,
  \item \textbf{coverage}: mean fraction of anchors present in the chosen output,
  \item \textbf{source coverage}: mean fraction of source phrases present in the chosen output,
  \item \textbf{ROUGE-L}: a reference-overlap proxy for surface quality,
  \item \textbf{token-$F_1$}: additional overlap signal,
  \item \textbf{runtime}: mean wall-clock latency per example.
\end{itemize}

The validation run covers 2{,}610 attainable constraint tasks: 993 from CommonGen, 996 from E2E NLG, and 621 from WikiBio. Each summary reports the sample mean and standard error over examples.

\Needspace{20\baselineskip}
\section{Experimental Setup}
All reported training and validation runs were executed on Apple Silicon using the Metal backend (\texttt{mps}). The hardware specification is reported only to contextualize wall-clock measurements. The main model was trained for several epochs with a batch size of 48 and a maximum sequence length of 160 tokens. The validation benchmark used:
\begin{itemize}[leftmargin=1.5em]
  \item beam size 6,
  \item best-of-16 ancestral sampling,
  \item 96 particles for twisted SMC,
  \item 64 maximum generated tokens,
  \item $\lambda=2.0$ and twist scale $\tau=2.0$,
  \item source-support scale $\beta=0.4$,
  \item ESS resampling threshold $0.5P$, splitting interval 12, and elite fraction 0.2.
\end{itemize}

\Needspace{22\baselineskip}
\section{Training Dynamics and Operating Frontier}
Table~\ref{tab:training-loss} reports the optimization trace for the prefix model. The loss trajectory is monotone over the reported epochs and supports the use of a fixed shared checkpoint for all decoding comparisons.

\begin{table}[H]
\centering
\caption{Epoch-wise optimization trace for the compact prefix model.}
\label{tab:training-loss}
\begin{tabular}{ccc}
\toprule
Epoch & Train loss & Validation loss \\
\midrule
1 & 4.574 & 3.992 \\
2 & 3.577 & 3.676 \\
3 & 3.263 & 3.498 \\
4 & 3.064 & 3.386 \\
5 & 2.917 & 3.326 \\
\bottomrule
\end{tabular}
\end{table}

Across datasets, the shared proposal model often realizes individual anchors without reliably realizing several anchors jointly under standard decoding. The comparison isolates how the inference procedure changes the operating point of a fixed conditional model.

\begin{figure}[H]
  \centering
  \includegraphics[width=\textwidth]{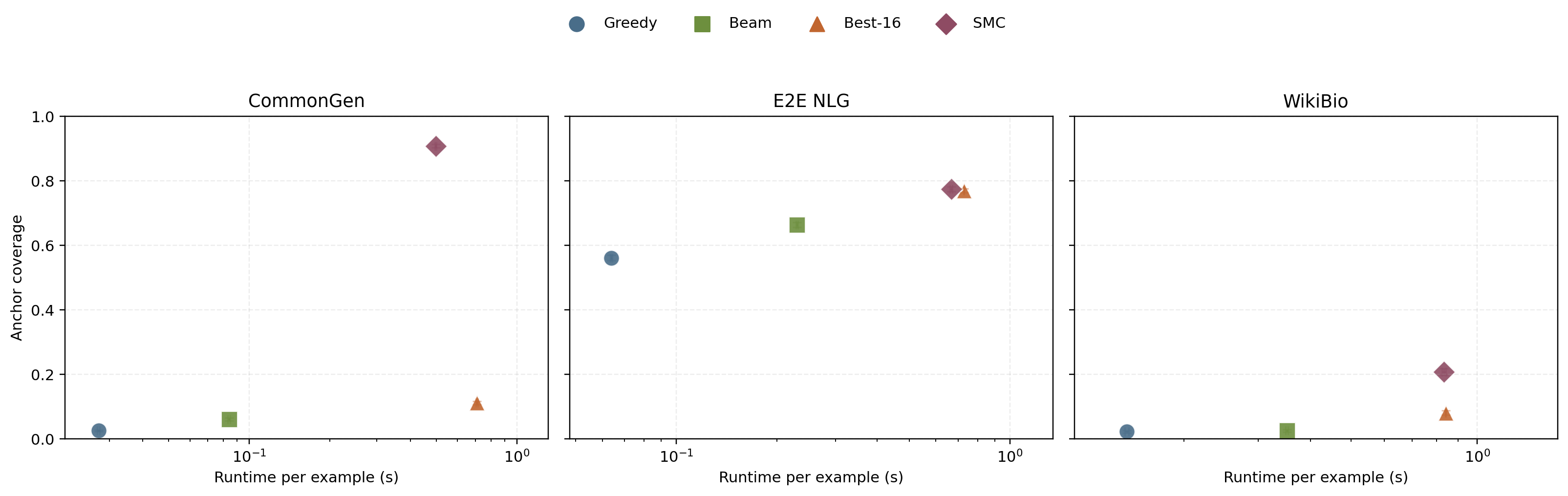}
  \caption{Coverage-runtime frontier for the validation benchmark. The x-axis is logarithmic to separate low-latency baselines from particle-based decoding regimes.}
  \label{fig:coverage-runtime}
\end{figure}

Figure~\ref{fig:coverage-runtime} gives a budget-oriented summary. On CommonGen, twisted SMC obtains substantially higher coverage and exact success than the tested baselines while remaining below the best-of-16 ancestral runtime. On E2E, twisted SMC improves exact success and coverage over greedy and beam filtering, and it obtains higher coverage than best-of-16 ancestral sampling at lower mean latency. On WikiBio, exact acceptance remains rare for all methods, but twisted SMC still produces the widest coverage margin. The figure is therefore read as a constrained-inference operating curve rather than as a single-metric ranking.

\Needspace{18\baselineskip}
\noindent The frontier plot summarizes the comparison expanded in Table~\ref{tab:main-results}: particle control shifts the coverage--runtime operating point, while overlap is measured separately to avoid conflating constraint satisfaction with surface similarity.
\FloatBarrier

\section{Results on Structured Generation}
\subsection{CommonGen}
Table~\ref{tab:main-results} shows the main validation result. On CommonGen, greedy, beam-filtered, and best-of-16 ancestral decoding do not produce exact three-anchor successes in the reported validation set. Twisted SMC reaches $0.758\pm0.014$ exact success, $0.908\pm0.006$ required-anchor coverage, and $0.772\pm0.007$ source coverage.

\subsection{E2E NLG}
On E2E, the base model already copies a nontrivial fraction of attributes under greedy decoding. Twisted SMC improves exact success to $0.473\pm0.016$ and required-anchor coverage to $0.774\pm0.008$, exceeding greedy, beam filtering, and best-of-16 ancestral sampling while maintaining lower mean latency than best-of-16 ancestral sampling. Source coverage is lower than the best-of-16 ancestral baseline, but source intrusion is also lower ($0.926$ versus $1.362$), which indicates that exact satisfaction, full source preservation, and unsupported-value avoidance are related but distinct criteria.

\subsection{WikiBio}
WikiBio introduces longer source serializations and more heterogeneous attribute values than the other two datasets. In this regime, exact satisfaction is difficult for every decoder, but twisted SMC raises exact success to $0.023\pm0.006$ and required-anchor coverage to $0.207\pm0.011$, compared with near-zero exact success for the baselines.

\begin{table}[H]
\centering
\footnotesize
\caption{Validation benchmark over 2{,}610 attainable constraint tasks. Parenthesized values are standard errors. Bold marks the best success, required coverage, and source coverage within each dataset.}
\label{tab:main-results}
\begin{tabular}{llccccc}
\toprule
Dataset & Method & Success & ReqCov & SrcCov & ROUGE-L & Runtime (s) \\
\midrule
\multirow{4}{*}{CommonGen}
& Greedy & 0.000 (0.000) & 0.025 (0.003) & 0.029 (0.003) & 0.222 (0.003) & 0.027 \\
& Beam filter & 0.000 (0.000) & 0.060 (0.004) & 0.072 (0.004) & 0.192 (0.003) & 0.084 \\
& Best-of-16 ancestral & 0.000 (0.000) & 0.111 (0.005) & 0.122 (0.005) & 0.173 (0.003) & 0.709 \\
& Twisted SMC & \textbf{0.758} (0.014) & \textbf{0.908} (0.006) & \textbf{0.772} (0.007) & 0.240 (0.003) & 0.498 \\
\midrule
\multirow{4}{*}{E2E NLG}
& Greedy & 0.149 (0.011) & 0.561 (0.008) & 0.508 (0.005) & 0.434 (0.005) & 0.064 \\
& Beam filter & 0.252 (0.014) & 0.664 (0.008) & 0.604 (0.005) & 0.455 (0.005) & 0.230 \\
& Best-of-16 ancestral & 0.391 (0.015) & 0.769 (0.006) & \textbf{0.628} (0.004) & 0.370 (0.004) & 0.728 \\
& Twisted SMC & \textbf{0.473} (0.016) & \textbf{0.774} (0.008) & 0.525 (0.007) & 0.286 (0.003) & 0.667 \\
\midrule
\multirow{4}{*}{WikiBio}
& Greedy & 0.000 (0.000) & 0.022 (0.004) & 0.017 (0.002) & 0.201 (0.004) & 0.146 \\
& Beam filter & 0.000 (0.000) & 0.025 (0.004) & 0.023 (0.002) & 0.181 (0.004) & 0.352 \\
& Best-of-16 ancestral & 0.002 (0.002) & 0.081 (0.007) & 0.062 (0.003) & 0.165 (0.003) & 0.841 \\
& Twisted SMC & \textbf{0.023} (0.006) & \textbf{0.207} (0.011) & \textbf{0.065} (0.003) & 0.128 (0.004) & 0.833 \\
\bottomrule
\end{tabular}
\end{table}

\FloatBarrier

\subsection{Interpretation}
The validation results support three observations:
\begin{enumerate}[leftmargin=1.5em]
  \item the proposal model can realize many anchors but does not reliably combine them under standard decoding,
  \item SMC reweighting, resampling, source-support proposal shaping, and splitting increase exact satisfaction and mean required coverage under the same checkpoint across three different structured-generation regimes,
  \item the bridge changes the operating point rather than dominating every metric: required-anchor coverage rises substantially, while source coverage, source intrusion, and ROUGE-L depend on how much of the structured input the baseline already copies and how often it substitutes unsupported values.
\end{enumerate}

\section{Source-Fidelity Diagnostics}
Exact realization of the designated anchor set is a necessary condition for the constrained evaluation target, but it is not a sufficient statistic for instance-level faithfulness. The accepting event is defined over a selected subset of source phrases, whereas the structured input may contain additional values that can be contradicted or replaced by unsupported alternatives. We report a complementary source-coverage statistic over the full phrase inventory $\mathcal{P}(x)$ exposed by the adapter:
\[
\mathrm{SCov}(y;x)
=
\frac{1}{|\mathcal{P}(x)|}
\sum_{c \in \mathcal{P}(x)}
\mathbf{1}\{c \text{ appears in } y\}.
\]
This statistic is not used to define attainability, since many legitimate references verbalize only part of a source record. It serves as an auxiliary diagnostic for value preservation under the same generated surface. It is stricter than required-anchor coverage in the sense that it evaluates the complete source phrase set rather than only the subset used to construct the rare accepting event.

Source coverage is paired with a source-intrusion statistic. Let $\mathcal{U}_{\mathcal{D}}$ denote the content-term vocabulary induced by candidate phrases in the evaluation subset, after removing terms whose document frequency exceeds a corpus-level high-frequency threshold, and let $\mathcal{U}(x)$ denote the corresponding term set for the current source instance. The intrusion count is
\[
\mathrm{Intr}(y;x)
=
\sum_{u \in \mathcal{U}_{\mathcal{D}}\setminus \mathcal{U}(x)}
\mathbf{1}\{u \text{ appears in } y\}.
\]
This quantity detects cases in which a decoder realizes the required anchors while importing lexical evidence supported by another source instance. Among continuations with the same exact-acceptance status, the implementation ranks required coverage first, source coverage second, source intrusion third, and model score plus reference overlap afterward. This ordering preserves the constrained evaluation target while penalizing unsupported value substitutions.

In the validation run, twisted SMC has lower mean source intrusion than the tested baselines on all three datasets: $0.226$ on CommonGen, $0.926$ on E2E NLG, and $6.332$ on WikiBio. The E2E result is particularly informative because exact success and source coverage disagree: best-of-16 ancestral sampling has the highest source coverage, whereas twisted SMC has the highest exact success and the lowest unsupported-value intrusion.

\section{Qualitative Interpretation}
The output examples in Appendix~\ref{sec:example-diagnostics} illustrate the same control--quality relation measured in the aggregate metrics. Incomplete continuations can remain competitive under local likelihood, while the twisted bridge reallocates probability mass toward accepting sequences. The examples are reported with coverage, source coverage, source-intrusion count, acceptance mass, and overlap so that exact acceptance is not interpreted as semantic adequacy by itself.

\section{Connection to Doob Transforms and Dynamic Systems}
Twisted proposals can be derived from the Doob $h$-transform associated with the probability of future acceptance. For the sequence model considered here, the exact harmonic function is not available because it depends on the probability of reaching an accepting automaton state from every possible recurrent hidden state and partial surface. LatticeBridge therefore uses distance reduction in the surface automaton together with a source-support proposal factor as computable surrogates.

From a dynamic-systems perspective, the partial sequence and automaton state define a controlled nonlinear state process
\[
z_t = (h_t, s_t),
\]
where $h_t$ is the recurrent hidden state and $s_t$ is the constraint state. The twist then acts as an online control law over the proposal kernel:
\[
q_t(\cdot \mid z_t) = \mathcal{T}_{\tau,\beta}\!\left(p_\theta(\cdot \mid z_t), d(s_t), \psi_x\right).
\]
Under this interpretation, the reported diagnostics follow directly from the controlled particle dynamics. Effective sample size measures degeneracy in the weighted particle approximation, acceptance mass estimates how much probability remains on accepting states, and runtime quantifies the hardware cost of applying the control law. Accelerator-oriented work on automata and trie vectorization \citep{su2026vectorizing} addresses the systems side of constrained decoding, while language-model SMC work \citep{zhao2024probabilistic,wu2024step} provides the probabilistic background.

\section{Approximate Doob Control Analysis}
\subsection{Exact Doob recursion}
For a fixed horizon $T$, the variance-minimizing proposal for conditioning on acceptance is the Doob-transformed kernel induced by the future acceptance probability. Let
\[
z_t = (h_t, s_t)
\]
denote the recurrent hidden state and automaton state after step $t$. Define the harmonic bridge function
\[
H_t(z_t) = \mathbb{P}_{p_\theta}\!\left(A(y_{1:T};\mathcal{C}) = 1 \mid z_t\right),
\]
with terminal condition $H_T(z_T)=A(y_{1:T};\mathcal{C})$. If $F(z_t,y)$ denotes the next state after emitting token $y$, then $H_t$ obeys the backward recursion
\[
H_t(z_t) = \sum_{y} p_\theta(y \mid z_t)\, H_{t+1}(F(z_t,y)).
\]
The corresponding zero-variance proposal is
\[
q_t^\star(y \mid z_t)
=
p_\theta(y \mid z_t)
\frac{H_{t+1}(F(z_t,y))}{H_t(z_t)}.
\]
Evaluating this proposal would require a backward recursion over the product of the language-model state and the automaton state. The automaton component is finite, but the recurrent hidden state is continuous, so the exact recursion is not tractable in the present implementation.

\subsection{Distance-driven twist with source support}
The implemented proposal replaces the unknown harmonic bridge with a surrogate defined by automaton distance and source support. Writing the one-step progress as
\[
\Delta_t(y) = d(s_t) - d(F_s(s_t,y)),
\]
where $F_s$ is the automaton transition, the implementation uses
\[
q_t(y \mid z_t) \propto p_\theta(y \mid z_t)\exp\{\tau \Delta_t(y) + \beta \psi_x(y)\}.
\]
Let
\[
Z_t(z_t) = \sum_y p_\theta(y \mid z_t)\exp\{\tau \Delta_t(y) + \beta \psi_x(y)\}.
\]
Because the sequence target itself uses the potential $\exp\{\lambda \Delta_t\}$, the incremental importance correction simplifies to
\[
\log \frac{G_t(y)\,p_\theta(y \mid z_t)}{q_t(y \mid z_t)}
=
(\lambda-\tau)\Delta_t(y) - \beta \psi_x(y) + \log Z_t(z_t).
\]
This identity separates two sources of variance:
\begin{enumerate}[leftmargin=1.5em]
  \item the mismatch between the target bridge strength $\lambda$ and the proposal twist strength $\tau$,
  \item the mismatch between local distance reduction plus source support and the true future acceptance value encoded in $H_t$.
\end{enumerate}
In the validation benchmark $\tau=\lambda$, so the first term vanishes and the remaining correction reduces to the source-support compensation plus the proposal normalizer. The distance and support potentials are not estimates of the exact harmonic function; they are deterministic control signals over compiled source evidence.

\subsection{Control-budget coupling}
The coverage--latency relation can be formalized as a control problem with resource penalties. One diagnostic objective for this relation is
\[
\mathcal{J}
=
\mathbb{E}\!\left[\mathrm{cov}(y_{1:T})\right]
-\eta_{\mathrm{lat}}\,
\mathbb{E}\!\left[\mathrm{time}(y_{1:T})\right]
-\eta_{\mathrm{deg}}
\sum_{t=1}^{T}
\mathbb{E}\!\left[\frac{P}{\mathrm{ESS}_t}-1\right]_+,
\]
where $\mathrm{cov}(y_{1:T})$ is anchor coverage, $P$ is the particle count, and the last term penalizes degeneracy. The implementation does not optimize this objective directly; it is used to interpret coverage, latency, and ESS under a common resource-sensitive view.

This view also distinguishes resampling from splitting. Resampling redistributes probability mass when normalized weights degenerate. Splitting reallocates finite compute toward high-scoring branches when acceptance becomes too rare under the current proposal. Both operations act on the same finite particle budget, but they address different numerical pathologies.

\section{Related Work}
\subsection{Structured data-to-text generation}
CommonGen \citep{lin2020commongen}, E2E NLG \citep{novikova2017e2e,dusek2019semantic}, and WikiBio \citep{lebret2016generating} are canonical benchmarks for testing whether a system can convert structured inputs into faithful text. Much work in this space focuses on architecture and pretraining rather than on the inference problem that appears after a model has already been trained.

\subsection{Constrained decoding}
Lexically constrained beam search \citep{hokamp2017lexically,post2018fast}, automata-based constrained decoding \citep{koo2024automata}, lookahead-based constrained decoding \citep{lu2022neurologic}, search-oriented inference-time scaling such as A*-decoding \citep{chatziveroglou2025astar}, draft-conditioned or mixed natural/structured decoding \citep{reddy2026draft,nguyen2026thinking}, retrieval-constrained systems such as trie-based decoders \citep{su2026vectorizing}, and recent MCMC-based constrained sampling \citep{anaya2025constrained} all address subsets of the same issue: how to keep decoding inside a constrained region without destroying throughput. Our work differs in using a particle system that explicitly represents multiple partial trajectories and their weights.

\subsection{Controlled language generation}
Methods such as PPLM \citep{dathathri2020pplm} and DExperts \citep{liu2021dexperts} alter decoding through gradients, auxiliary models, or expert/anti-expert logits. These methods are powerful but often depend on extra control models or attribute classifiers. LatticeBridge instead relies on instance-provided anchors and automaton geometry.

\subsection{Sequential Monte Carlo and twisting}
The theoretical backbone comes from Feynman--Kac particle systems \citep{delmoral2004feynman}, sequential Monte Carlo samplers \citep{delmoral2006smc}, and modern tutorials such as \citet{naesseth2019elements}. Twisted particle filters and related methods \citep{whiteley2014twisted,guarniero2017iterated} motivate our use of future-oriented proposal shaping. Recent sequence-level extensions include twisted SMC for language-model inference \citep{zhao2024probabilistic}, reasoning-oriented twisted sampling \citep{wu2024step}, and 2026 work on language-model ensembling with shared character-space SMC \citep{chan2026ensembling}. Our implementation is closer in spirit to this language-model inference line than to classical state-space estimation, but the numerical concerns are the same.

\section{Control Parameters and Validation Checks}
The benchmark record stores the decoding controls used for every reported run. The validation configuration fixes the split, task count, maximum generation length, beam size, ancestral sample count, particle count, bridge strength, twist strength, source-support strength, ESS threshold, splitting interval, elite fraction, device, and random seed.

The same checkpoint and tokenizer are used by all four decoders. Candidate selection is performed after decoding by the ordered key
\[
(\mathbf{1}\{\text{all anchors realized}\},\ \mathrm{ReqCov},\ \mathrm{SCov},\ -\mathrm{Intr},\ \log p,\ \mathrm{ROUGE\mbox{-}L}),
\]
so exact satisfaction is primary, partial required coverage and source coverage are secondary, source-intrusion avoidance is tertiary, and model score plus reference overlap are used only to break remaining ties. This policy is applied uniformly to beam filtering, ancestral sampling, and SMC outputs.

For the SMC decoder, the validation run also records mean ESS, resampling count, acceptance mass, and realized generation length. On CommonGen, the mean ESS is 57.07 for 96 particles and the mean acceptance mass is 0.782. On E2E NLG, the mean ESS is 56.25 and the acceptance mass is 0.186. On WikiBio, the mean ESS is 44.57 and the acceptance mass is 0.033, indicating the lowest accepting mass among the three reported datasets. These diagnostics report particle degeneracy directly rather than inferring it from coverage alone.

\section{Conclusion}
LatticeBridge treats faithful structured sequence synthesis as sequential rare-event inference. The implementation combines a compact prefix model, surface automata, support-aware twisted SMC, and multilevel splitting in a form that exposes the control mechanism through recorded particle statistics. In the multi-dataset validation benchmark, the bridge improves exact satisfaction and required-anchor coverage relative to standard decoders while reporting the associated source-coverage, source-intrusion, overlap, and latency measurements.

The system provides a reusable automaton interface, particle diagnostics, and benchmark files for studying the coverage--fidelity--runtime operating surface under a fixed proposal model.

\appendix

\section{Implementation Details}
\subsection{Tokenizer}
We use a byte-level BPE tokenizer with special tokens for source and target delimiters. Byte-level modeling keeps the surface automata general enough to handle punctuation and mixed-case entity names without introducing language-specific preprocessing.

\subsection{Model}
The current checkpoint uses a 256-dimensional embedding, a 384-dimensional GRU hidden state, two recurrent layers, and dropout of 0.15. The objective is masked next-token prediction over the target portion only.

\subsection{Inference}
Greedy, beam-filtered, best-of-$k$, and twisted SMC all share the same prefix model. The comparison therefore isolates the inference procedure rather than conflating decoding with model scale or pretraining.

\section{Evaluation Scope}
The evaluation is defined with respect to exact realization of input-derived anchors. This choice yields an explicit accepting event together with a transition system over compiled surface automata. The reported results should be interpreted as measurements of anchor-faithful generation rather than as claims about broader semantic adequacy.

The abstraction supplies an acceptance lattice that is shared across datasets while keeping schema-specific processing outside the inference core.

\section{Budget-Normalized Diagnostics}
Table~\ref{tab:efficiency-deltas} reports coverage and success lift relative to greedy decoding, normalized by additional runtime. On CommonGen, twisted SMC is the only tested method that reaches exact success. On E2E, best-of-16 ancestral sampling and twisted SMC both improve exact success, while twisted SMC gives the larger success lift at lower mean latency. On WikiBio, exact gains remain small, but the coverage lift of twisted SMC is larger than that of the baselines.

\begin{table}[H]
\centering
\caption{Budget-normalized lift relative to greedy decoding. Positive values indicate improvement beyond the greedy baseline.}
\label{tab:efficiency-deltas}
\begin{tabular}{llcccc}
\toprule
Dataset & Method & $\Delta$ Success & $\Delta$ Coverage & Cov./extra s & Succ./extra s \\
\midrule
\multirow{3}{*}{CommonGen}
& Beam filter & 0.000 & 0.035 & 0.62 & 0.00 \\
& Best-of-16 ancestral & 0.000 & 0.087 & 0.13 & 0.00 \\
& Twisted SMC & 0.758 & 0.883 & 1.88 & 1.61 \\
\midrule
\multirow{3}{*}{E2E NLG}
& Beam filter & 0.103 & 0.103 & 0.63 & 0.63 \\
& Best-of-16 ancestral & 0.242 & 0.208 & 0.31 & 0.36 \\
& Twisted SMC & 0.324 & 0.213 & 0.35 & 0.54 \\
\midrule
\multirow{3}{*}{WikiBio}
& Beam filter & 0.000 & 0.003 & 0.01 & 0.00 \\
& Best-of-16 ancestral & 0.002 & 0.059 & 0.08 & 0.00 \\
& Twisted SMC & 0.023 & 0.185 & 0.27 & 0.03 \\
\bottomrule
\end{tabular}
\end{table}

\section{Example Diagnostics}
\label{sec:example-diagnostics}
\begingroup
\scriptsize
\setlength{\tabcolsep}{2pt}
\setlength{\LTleft}{0pt}
\setlength{\LTright}{0pt}
\renewcommand{\arraystretch}{1.08}

\subsection*{CommonGen}
\begin{longtable}{>{\raggedright\arraybackslash}p{0.10\textwidth}>{\raggedright\arraybackslash}p{0.24\textwidth}>{\raggedright\arraybackslash}p{0.22\textwidth}>{\raggedright\arraybackslash}p{0.22\textwidth}>{\raggedright\arraybackslash}p{0.11\textwidth}}
\toprule
Case & Anchors & Best baseline & LatticeBridge & Particle state \\
\midrule
\endfirsthead
\toprule
Case & Anchors & Best baseline & LatticeBridge & Particle state \\
\midrule
\endhead
exact lift & house, table, sit & beam\_filter; req 0.67; src 0.67; intr 0; RL 0.40 & req 1.00; src 1.00; intr 0; RL 0.67 & mass 1.00; ESS 54.6 \\
\midrule
exact lift & smoke, blow, sit & ancestral\_best\_of\_k; req 0.33; src 0.33; intr 0; RL 0.09 & req 1.00; src 1.00; intr 0; RL 0.29 & mass 1.00; ESS 53.6 \\
\midrule
exact lift & mirror, shave, face & ancestral\_best\_of\_k; req 0.33; src 0.25; intr 0; RL 0.22 & req 1.00; src 0.75; intr 0; RL 0.40 & mass 1.00; ESS 59.7 \\
\midrule
exact lift & snowball, snow, kid & beam\_filter; req 0.00; src 0.00; intr 1; RL 0.23 & req 1.00; src 0.75; intr 0; RL 0.22 & mass 1.00; ESS 56.7 \\
\midrule
coverage lift & pierce, ear, chair & beam\_filter; req 0.00; src 0.00; intr 1; RL 0.32 & req 0.67; src 0.50; intr 0; RL 0.35 & mass 0.00; ESS 43.9 \\
\midrule
coverage lift & sail, day, boat & ancestral\_best\_of\_k; req 0.33; src 0.33; intr 0; RL 0.21 & req 0.67; src 0.67; intr 0; RL 0.22 & mass 0.00; ESS 48.1 \\
\midrule
coverage lift & sleigh, pull, dog & ancestral\_best\_of\_k; req 0.33; src 0.33; intr 0; RL 0.24 & req 0.67; src 0.67; intr 0; RL 0.43 & mass 1.00; ESS 54.1 \\
\midrule
coverage lift & slide, hill, kid & beam\_filter; req 0.00; src 0.00; intr 1; RL 0.36 & req 0.67; src 0.67; intr 0; RL 0.50 & mass 0.00; ESS 36.8 \\
\midrule
\bottomrule
\end{longtable}

\subsection*{E2E NLG}
\begin{longtable}{>{\raggedright\arraybackslash}p{0.10\textwidth}>{\raggedright\arraybackslash}p{0.24\textwidth}>{\raggedright\arraybackslash}p{0.22\textwidth}>{\raggedright\arraybackslash}p{0.22\textwidth}>{\raggedright\arraybackslash}p{0.11\textwidth}}
\toprule
Case & Anchors & Best baseline & LatticeBridge & Particle state \\
\midrule
\endfirsthead
\toprule
Case & Anchors & Best baseline & LatticeBridge & Particle state \\
\midrule
\endhead
low-mass success & Cotto, moderate, The Portland Arms & ancestral\_best\_of\_k; req 1.00; src 1.00; intr 0; RL 0.44 & req 1.00; src 1.00; intr 0; RL 0.65 & mass 0.00; ESS 49.0 \\
\midrule
exact lift & 5 out of 5, cheap, Fitzbillies & ancestral\_best\_of\_k; req 0.67; src 0.50; intr 4; RL 0.45 & req 1.00; src 0.83; intr 0; RL 0.56 & mass 0.00; ESS 68.7 \\
\midrule
exact lift & Aromi, 5 out of 5, English & greedy; req 0.67; src 0.50; intr 2; RL 0.67 & req 1.00; src 0.83; intr 0; RL 0.56 & mass 0.00; ESS 62.0 \\
\midrule
exact lift & Aromi, restaurant, cheap & beam\_filter; req 0.67; src 0.75; intr 0; RL 0.57 & req 1.00; src 1.00; intr 0; RL 0.36 & mass 1.00; ESS 84.4 \\
\midrule
exact lift & Aromi, cheap, coffee shop & beam\_filter; req 0.67; src 0.67; intr 1; RL 0.40 & req 1.00; src 1.00; intr 0; RL 0.40 & mass 0.00; ESS 39.1 \\
\midrule
low-mass success & Cocum, 1 out of 5, high & beam\_filter; req 1.00; src 0.83; intr 1; RL 0.60 & req 1.00; src 0.83; intr 0; RL 0.54 & mass 0.00; ESS 73.6 \\
\midrule
low-mass success & Cotto, The Portland Arms, cheap & ancestral\_best\_of\_k; req 1.00; src 0.80; intr 1; RL 0.42 & req 1.00; src 1.00; intr 0; RL 0.40 & mass 0.00; ESS 44.7 \\
\midrule
low-mass success & Aromi, average, coffee shop & beam\_filter; req 1.00; src 1.00; intr 0; RL 0.53 & req 1.00; src 1.00; intr 0; RL 0.27 & mass 0.00; ESS 31.3 \\
\midrule
\bottomrule
\end{longtable}

\subsection*{WikiBio}
\begin{longtable}{>{\raggedright\arraybackslash}p{0.10\textwidth}>{\raggedright\arraybackslash}p{0.24\textwidth}>{\raggedright\arraybackslash}p{0.22\textwidth}>{\raggedright\arraybackslash}p{0.22\textwidth}>{\raggedright\arraybackslash}p{0.11\textwidth}}
\toprule
Case & Anchors & Best baseline & LatticeBridge & Particle state \\
\midrule
\endfirsthead
\toprule
Case & Anchors & Best baseline & LatticeBridge & Particle state \\
\midrule
\endhead
coverage lift & 1779, thomas edwards, welsh & beam\_filter; req 0.00; src 0.00; intr 2; RL 0.38 & req 0.67; src 0.25; intr 0; RL 0.52 & mass 0.00; ESS 50.0 \\
\midrule
coverage lift & alberto volpi, road & beam\_filter; req 0.00; src 0.00; intr 2; RL 0.25 & req 0.50; src 0.10; intr 0; RL 0.05 & mass 0.00; ESS 25.6 \\
\midrule
coverage lift & activist, joseph defilippis & ancestral\_best\_of\_k; req 0.00; src 0.10; intr 2; RL 0.27 & req 0.50; src 0.10; intr 0; RL 0.04 & mass 0.00; ESS 34.3 \\
\midrule
near miss & baltimore orioles, pitcher & ancestral\_best\_of\_k; req 0.50; src 0.07; intr 1; RL 0.33 & req 0.50; src 0.07; intr 0; RL 0.15 & mass 0.00; ESS 22.9 \\
\midrule
coverage lift & mason cook darling, may 18 , 1801, wisconsin & ancestral\_best\_of\_k; req 0.00; src 0.00; intr 2; RL 0.40 & req 0.33; src 0.08; intr 0; RL 0.35 & mass 0.00; ESS 46.0 \\
\midrule
near miss & theodore von eltz, actor & ancestral\_best\_of\_k; req 0.50; src 0.10; intr 3; RL 0.26 & req 0.50; src 0.10; intr 0; RL 0.05 & mass 0.00; ESS 50.1 \\
\midrule
near miss & christine isobel mcgaffey frederick, american & ancestral\_best\_of\_k; req 0.50; src 0.12; intr 3; RL 0.19 & req 0.50; src 0.12; intr 0; RL 0.02 & mass 0.00; ESS 48.7 \\
\midrule
near miss & barbara coombs lee, president of compassion \& choices, american & ancestral\_best\_of\_k; req 0.33; src 0.17; intr 2; RL 0.04 & req 0.33; src 0.17; intr 0; RL 0.01 & mass 0.00; ESS 74.0 \\
\midrule
\bottomrule
\end{longtable}
\endgroup

The examples report the same measurements used in Table~\ref{tab:main-results}: exact satisfaction, source coverage, source intrusion, overlap, acceptance mass, and particle behavior are interpreted jointly rather than as substitutable quality scores.


\begin{thebibliography}{27}
\providecommand{\natexlab}[1]{#1}
\providecommand{\url}[1]{\texttt{#1}}
\expandafter\ifx\csname urlstyle\endcsname\relax
  \providecommand{\doi}[1]{doi: #1}\else
  \providecommand{\doi}{doi: \begingroup \urlstyle{rm}\Url}\fi

\bibitem[Anaya~Gonzalez et~al.(2025)Anaya~Gonzalez, Vaidya, Park, Ji,
  Berg-Kirkpatrick, and D'Antoni]{anaya2025constrained}
Emmanuel Anaya~Gonzalez, Sairam Vaidya, Kanghee Park, Ruyi Ji, Taylor
  Berg-Kirkpatrick, and Loris D'Antoni.
\newblock Constrained sampling for language models should be easy: An mcmc
  perspective.
\newblock \emph{arXiv preprint arXiv:2506.05754}, 2025.

\bibitem[Cerou and Guyader(2007)]{cerou2007adaptive}
Frederic Cerou and Arnaud Guyader.
\newblock Adaptive multilevel splitting for rare event analysis.
\newblock \emph{Stochastic Analysis and Applications}, 25\penalty0
  (2):\penalty0 417--443, 2007.

\bibitem[Chan et~al.(2026)Chan, Liu, Kiegeland, Pasti, Vigly, O'Donnell,
  Cotterell, and Vieira]{chan2026ensembling}
Robin Shing~Moon Chan, Tianyu Liu, Samuel Kiegeland, Clemente Pasti,
  Jacob~Hoover Vigly, Timothy~J. O'Donnell, Ryan Cotterell, and Tim Vieira.
\newblock Ensembling language models with sequential monte carlo.
\newblock \emph{arXiv preprint arXiv:2603.05432}, 2026.

\bibitem[Chatziveroglou(2025)]{chatziveroglou2025astar}
Giannis Chatziveroglou.
\newblock A*-decoding: Token-efficient inference scaling.
\newblock \emph{arXiv preprint arXiv:2505.13672}, 2025.

\bibitem[Dathathri et~al.(2020)Dathathri, Madotto, Lan, Hung, Frank, Molino,
  Yosinski, and Liu]{dathathri2020pplm}
Sumanth Dathathri, Andrea Madotto, Janice Lan, Jane Hung, Eric Frank, Piero
  Molino, Jason Yosinski, and Rosanne Liu.
\newblock Plug and play language models: A simple approach to controlled text
  generation.
\newblock In \emph{Proceedings of ICLR}, 2020.

\bibitem[Del~Moral(2004)]{delmoral2004feynman}
Pierre Del~Moral.
\newblock \emph{Feynman-Kac Formulae: Genealogical and Interacting Particle
  Systems with Applications}.
\newblock Springer, 2004.

\bibitem[Del~Moral et~al.(2006)Del~Moral, Doucet, and Jasra]{delmoral2006smc}
Pierre Del~Moral, Arnaud Doucet, and Ajay Jasra.
\newblock Sequential monte carlo samplers.
\newblock \emph{Journal of the Royal Statistical Society: Series B},
  68\penalty0 (3):\penalty0 411--436, 2006.

\bibitem[Dusek et~al.(2019)Dusek, Howcroft, and Rieser]{dusek2019semantic}
Ondrej Dusek, David~M. Howcroft, and Verena Rieser.
\newblock Semantic noise matters for neural natural language generation.
\newblock In \emph{Proceedings of the 12th International Conference on Natural
  Language Generation}, pages 421--426, 2019.

\bibitem[Gehrmann et~al.(2021)Gehrmann, Adewumi, Aggarwal, Ammanamanchi,
  Anuoluwapo, Bosselut, Chandu, Clinciu, Das, Dhole, et~al.]{gehrmann2021gem}
Sebastian Gehrmann, Tosin Adewumi, Karmanya Aggarwal, Pawan~Sasanka
  Ammanamanchi, Aremu Anuoluwapo, Antoine Bosselut, Khyathi~Raghavi Chandu,
  Miruna Clinciu, Dipanjan Das, Kaustubh Dhole, et~al.
\newblock The gem benchmark: Natural language generation, its evaluation and
  metrics.
\newblock In \emph{Proceedings of the 1st Workshop on Natural Language
  Generation, Evaluation, and Metrics}, pages 96--120, 2021.

\bibitem[Guarniero et~al.(2017)Guarniero, Johansen, and
  Lee]{guarniero2017iterated}
Peter Guarniero, Adam~M. Johansen, and Anthony Lee.
\newblock The iterated auxiliary particle filter.
\newblock \emph{Journal of the American Statistical Association}, 112\penalty0
  (520):\penalty0 1636--1647, 2017.

\bibitem[Hokamp and Liu(2017)]{hokamp2017lexically}
Chris Hokamp and Qun Liu.
\newblock Lexically constrained decoding for sequence generation using grid
  beam search.
\newblock In \emph{Proceedings of ACL}, pages 1535--1546, 2017.

\bibitem[Kim et~al.(2025)Kim, Nam, Park, and Lee]{selfdistilled2025twisted}
Sooyeon Kim, Giung Nam, Byoungwoo Park, and Juho Lee.
\newblock Improving constrained language generation via self-distilled twisted
  sequential monte carlo.
\newblock \emph{arXiv preprint arXiv:2507.02315}, 2025.

\bibitem[Koo et~al.(2024)Koo, Liu, and He]{koo2024automata}
Terry Koo, Frederick Liu, and Luheng He.
\newblock Automata-based constraints for language model decoding.
\newblock \emph{arXiv preprint arXiv:2407.08103}, 2024.

\bibitem[Lebret et~al.(2016)Lebret, Grangier, and Auli]{lebret2016generating}
Remi Lebret, David Grangier, and Michael Auli.
\newblock Generating text from structured data with application to the
  biography domain.
\newblock \emph{arXiv preprint arXiv:1603.07771}, 2016.

\bibitem[Lhoest et~al.(2021)Lhoest, Villanova~del Moral, Jernite, Thakur, von
  Platen, Patil, Chaumond, Drame, Plu, Tunstall, et~al.]{lhoest2021datasets}
Quentin Lhoest, Albert Villanova~del Moral, Yacine Jernite, Abhishek Thakur,
  Patrick von Platen, Suraj Patil, Julien Chaumond, Mariama Drame, Julien Plu,
  Lewis Tunstall, et~al.
\newblock Datasets: A community library for natural language processing.
\newblock In \emph{Proceedings of the 2021 Conference on Empirical Methods in
  Natural Language Processing: System Demonstrations}, pages 175--184, 2021.

\bibitem[Lin et~al.(2020)Lin, Zhou, Shen, Zhou, Bhagavatula, Choi, and
  Ren]{lin2020commongen}
Bill~Yuchen Lin, Wangchunshu Zhou, Ming Shen, Pei Zhou, Chandra Bhagavatula,
  Yejin Choi, and Xiang Ren.
\newblock Commongen: A constrained text generation challenge for generative
  commonsense reasoning.
\newblock In \emph{Findings of the Association for Computational Linguistics:
  EMNLP 2020}, pages 1823--1840, 2020.

\bibitem[Liu et~al.(2021)Liu, Sap, Lu, Swayamdipta, Bhagavatula, Smith, and
  Choi]{liu2021dexperts}
Alisa Liu, Maarten Sap, Ximing Lu, Swabha Swayamdipta, Chandra Bhagavatula,
  Noah~A. Smith, and Yejin Choi.
\newblock Dexperts: Decoding-time controlled text generation with experts and
  anti-experts.
\newblock In \emph{Proceedings of ACL-IJCNLP}, pages 6691--6706, 2021.

\bibitem[Lu et~al.(2022)Lu, West, Zellers, Bras, Bhagavatula, and
  Choi]{lu2022neurologic}
Ximing Lu, Peter West, Rowan Zellers, Ronan~Le Bras, Chandra Bhagavatula, and
  Yejin Choi.
\newblock Neurologic a*esque decoding: Constrained text generation with
  lookahead heuristics.
\newblock In \emph{Proceedings of NAACL-HLT}, pages 780--799, 2022.

\bibitem[Naesseth et~al.(2019)Naesseth, Lindsten, and
  Schon]{naesseth2019elements}
Christian~A. Naesseth, Fredrik Lindsten, and Thomas~B. Schon.
\newblock Elements of sequential monte carlo.
\newblock \emph{Foundations and Trends in Machine Learning}, 12\penalty0
  (3):\penalty0 307--392, 2019.

\bibitem[Nguyen et~al.(2026)Nguyen, Silva, Zumot, Tupikina, Aghasaryan, and
  Alam]{nguyen2026thinking}
Ngoc Trinh~Hung Nguyen, Alonso Silva, Laith Zumot, Liubov Tupikina, Armen
  Aghasaryan, and Mehwish Alam.
\newblock Thinking before constraining: A unified decoding framework for large
  language models.
\newblock \emph{arXiv preprint arXiv:2601.07525}, 2026.

\bibitem[Novikova et~al.(2017)Novikova, Dusek, and Rieser]{novikova2017e2e}
Jekaterina Novikova, Ondrej Dusek, and Verena Rieser.
\newblock The e2e dataset: New challenges for end-to-end generation.
\newblock In \emph{Proceedings of SIGDIAL}, pages 201--206, 2017.

\bibitem[Post and Vilar(2018)]{post2018fast}
Matt Post and David Vilar.
\newblock Fast lexically constrained decoding with dynamic beam allocation for
  neural machine translation.
\newblock In \emph{Proceedings of NAACL-HLT}, pages 1314--1324, 2018.

\bibitem[Reddy et~al.(2026)Reddy, Walker, Ide, and Bedi]{reddy2026draft}
Avinash Reddy, Thayne~T. Walker, James~S. Ide, and Amrit~Singh Bedi.
\newblock Draft-conditioned constrained decoding for structured generation in
  llms.
\newblock \emph{arXiv preprint arXiv:2603.03305}, 2026.

\bibitem[Su et~al.(2026)Su, Katsman, Wang, He, Heldt, Keshavan, Wang, Yi, Gao,
  Dalal, Hong, Chi, and Han]{su2026vectorizing}
Zhengyang Su, Isay Katsman, Yueqi Wang, Ruining He, Lukasz Heldt, Raghunandan
  Keshavan, Shao-Chuan Wang, Xinyang Yi, Mingyan Gao, Onkar Dalal, Lichan Hong,
  Ed~Chi, and Ningren Han.
\newblock Vectorizing the trie: Efficient constrained decoding for llm-based
  generative retrieval on accelerators.
\newblock \emph{arXiv preprint arXiv:2602.22647}, 2026.

\bibitem[Whiteley et~al.(2016)Whiteley, Lee, and Heine]{whiteley2014twisted}
Nick Whiteley, Anthony Lee, and Karolina Heine.
\newblock Twisted particle filters.
\newblock \emph{The Annals of Statistics}, 44\penalty0 (2):\penalty0 822--859,
  2016.

\bibitem[Wu et~al.(2024)Wu, Yao, Chen, Pan, Wang, Liu, and Yu]{wu2024step}
Xuansheng Wu, Wenlin Yao, Jianshu Chen, Xiaoman Pan, Xiaoyang Wang, Ninghao
  Liu, and Dong Yu.
\newblock Step-by-step reasoning for math problems via twisted sequential monte
  carlo.
\newblock \emph{arXiv preprint arXiv:2410.01920}, 2024.

\bibitem[Zhao et~al.(2024)Zhao, Brekelmans, Makhzani, and
  Grosse]{zhao2024probabilistic}
Stephen Zhao, Rob Brekelmans, Alireza Makhzani, and Roger~B. Grosse.
\newblock Probabilistic inference in language models via twisted sequential
  monte carlo.
\newblock In \emph{Proceedings of the 41st International Conference on Machine
  Learning}, pages 60704--60748, 2024.

\end{thebibliography}
\end{document}